\documentclass{article}
\usepackage{spconf,amsmath,graphicx,hyperref}
\usepackage{amssymb}
\usepackage{booktabs}

\usepackage{multirow} 

\title{AnimateScene: Camera-controllable Animation in Any Scene}
\name{
  \begin{tabular}{c}
    Qingyang Liu$^{1}$, Bingjie Gao$^{1}$, Weiheng Huang$^{2}$, Jun Zhang$^{2}$, Zhongqian Sun$^{2}$, Yang Wei$^{2}$, Fengrui Liu$^{3}$, \\
    Zelin Peng$^{1}$, Qianli Ma$^{1}$, Shuai Yang$^{1}$, Zhaohe Liao$^{1}$, Haonan Zhao$^{1}$, Li Niu$^{1}$\sthanks{Corresponding author.}
  \end{tabular}
}
\address{
$^{1}$ Shanghai Jiao Tong University, Shanghai, China\\
$^{2}$ Tencent TEG, Shenzhen, China\\
$^{3}$ East China Normal University, Shanghai, China
}
\begin{document}
%
\maketitle
\begin{abstract}
Recent advances in 3D scene reconstruction and 4D human animation have broadened adoption, but integrating the two remains difficult. Key challenges include placing humans at plausible locations and scales without interpenetration, aligning lighting and style between humans and background, and handling dynamic camera trajectories. We introduce AnimateScene, a unified framework addressing these issues. First, an accurate placement module automatically determines realistic 3D positions and prevents collisions during motion. Second, we propose a training-free style alignment method that adapts the 4D human to match background illumination and appearance, achieving coherent composites. Finally, we design a joint post-reconstruction approach that incorporates camera trajectories for smooth, visually engaging motion videos. Experiments across diverse scenes and actions demonstrate that AnimateScene generates dynamic results with high geometric detail and strong spatiotemporal coherence.  The project page is available at \href{https://whynothaha.github.io/AnimateScene/AnimateScene.html}{this https URL}.

\end{abstract}
\section{Introduction}
Camera-controllable animation aims to build dynamic scenes consistent with both subject motion and camera shifts. Recent work has explored adding either camera control~\cite{liu2024splatraj, liu2024gs} or human motion~\cite{liu2021neural, gao2023neural} into the 3D Gaussian Splatting (3DGS)\cite{kerbl20233d} framework, but a unified approach combining both is still missing. The goal is to generate vivid 4D human sequences with controllable camera trajectories from a single scene image and a single human image. Although video generation methods have attempted similar tasks, they suffer from slow inference and lack of explicit 3D constraints.

A straightforward solution is to combine single-view 3D scene reconstruction~\cite{che2024gamegen,zhao2024genxd} with animatable 4D human motion~\cite{yu2024wonderworld}. However, this raises challenges: (1) correctly positioning and scaling the human without unrealistic overlap; (2) handling mismatched lighting or style between foreground and background~\cite{feng2024crestyler}; and (3) ensuring smooth, coherent camera trajectories, which requires reconstructing viewpoints along a defined path. These issues make seamless scene–character integration particularly difficult.

To address these challenges, we propose AnimateScene. For placement, it introduces a depth-guided 3D module that infers collision-free locations and scales for the human avatar. A temporal smoother maintains stable positioning across the motion sequence. For style alignment, a training-free transfer module injects scene lighting and color into the human while preserving geometry, yielding visually coherent composites. For camera control, a joint post-reconstruction module fuses the human and scene Gaussian fields and fills gaps with diffusion-based inpainting, ensuring artifact-free renders along the entire camera trajectory.

AnimateScene requires only four inputs: a scene image, a human image with motion, a motion clip, and a user-defined camera path. It harmonizes appearance, reconstructs animatable 4D human avatars and sparse 3D backgrounds, performs depth-guided placement, and finally refines the fused human–scene field with post-reconstruction inpainting. Extensive experiments across diverse indoor and outdoor environments show that AnimateScene consistently outperforms state-of-the-art baselines in rendering quality, geometric consistency, and placement plausibility, demonstrating strong generalization to dynamic 4D scenarios.

In summary, our main contributions are as follows: (1) We introduce AnimateScene, a unified framework that couples single-image scene reconstruction, style-aligned 4D humans and controllable camera paths, enabling joint control over background and actor motion. (2) We propose a depth-guided 3D placement module that upgrades 2D object-placement predictions into collision-free world coordinates, guaranteeing physically coherent integration of animatable 3D Gaussian avatars. (3) We propose a joint post-reconstruction module to eliminate foreground-background occlusion, and ensure geometric and stylistic consistency across varied camera–action combinations.

\section{Method}
\label{sec:method}

\begin{figure*}[tb]
    \centering
\includegraphics[width=1.0\textwidth]{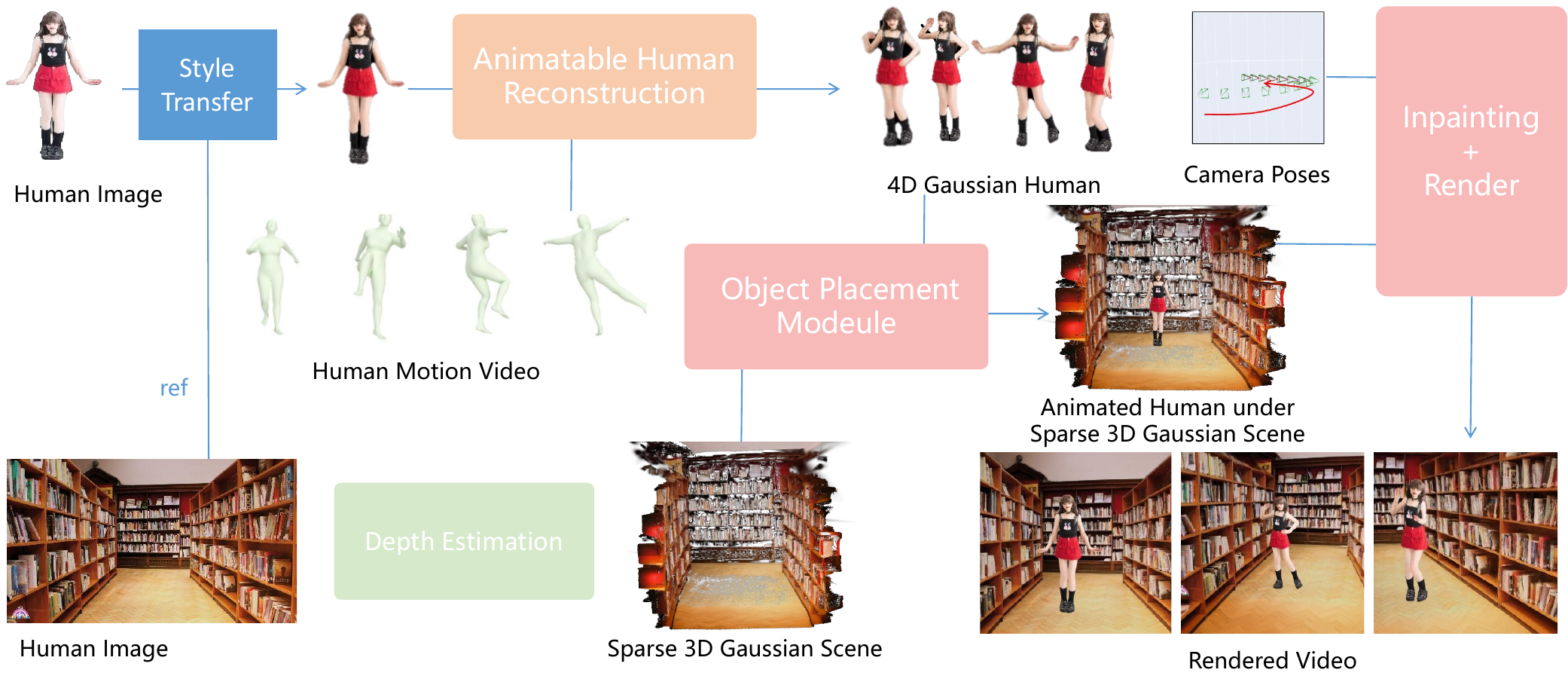}
    \caption{\textbf{Overview of AnimateScene.} AnimateScene takes a single scene image, a single human image, an accompanying motion clip, and a user-defined camera path. It first aligns the human’s appearance with the scene. Next, it reconstructs a 4D Gaussian avatar alongside a sparse 3D Gaussian scene. Depth cues are then used to lift the avatar to a collision-free 3D location and scale. The system finally refines the fused human–scene field along the camera trajectory, inpainting any newly exposed regions. The result is a video where actor, environment, and viewpoint move in seamless geometric and stylistic harmony.}
    \label{fig:overview}
\end{figure*}

As illustrated in Fig.~\ref{fig:overview}, AnimateScene consists of three main parts, 1) an \textit{animatable human style transfer module}, 2) an \textit{object placement model}, as well as 3) a \textit{trajectory‑based joint human–scene reconstruction model}. Given an input RGB human image \( I_h \in \mathbb{R}^{H \times W \times 3} \), an input RGB scene image \( I_s \in \mathbb{R}^{H \times W \times 3} \), a target camera trajectory $\boldsymbol{T}=\left\{\boldsymbol{T}_{\mathrm{i}}\right\}_{\mathrm{i}=1}^{\mathrm{n}} \in \mathbb{R}^{\mathrm{n} \times 4 \times 4}$, and a human motion video \( V_h \in \mathbb{R}^{F \times H \times W \times 3} \), we first use the style transfer module to align the style of the human image $I_h$ with that of the scene image $I_s$. Then the animatable human reconstruction model can reconstruct an style-consistent animatable 4D human avatar sequence. Single-view scene reconstruction model is utilized to reconstruct a 3D scene from a single-view image \(I_s\). The object placement model predicts a reasonable placement for \(I_h\) in terms of \(I_s\) , involving location \((x,y,z)\) and scale \((w,h)\).  The animated 3D gaussian human is placed in the reconstructed 3D scene with the guidance of the predicted placement. Finally, we jointly reconstruct the fused 3D scene along the camera trajectory $\boldsymbol{T}$.

\subsection{Preliminaries}
\noindent \textbf{3D Gaussian Splatting.} 3D Gaussian Splatting~\cite{kerbl20233d, he2025see} represents a scene as a collection of anisotropic Gaussian primitives, each defined by a centroid $\mathbf{x}\in\mathbb{R}^3$, scale parameters $\mathbf{s}\in\mathbb{R}^3$, and a rotation quaternion $\mathbf{q}\in\mathbb{R}^4$, together with an opacity $\alpha\in\mathbb{R}$ and an appearance feature vector $\mathbf{c}\in\mathbb{R}^C$. View-dependent shading is captured by encoding $\mathbf{c}$ with spherical harmonics. During rendering, each 3D Gaussian is projected into the image plane as a 2D Gaussian and composited front-to-back via alpha blending.

\noindent \textbf{Animatable Human Reconstruction.}
We use SMPL/ SMPL-X~\cite{loper2015smpl} to parameterize a human mesh with shape $\boldsymbol{\beta}\in\mathbb{R}^{20}$ and pose $\boldsymbol{\theta}\in\mathbb{R}^{55\times3}$, yielding a compact, deformable articulation model. For animatable reconstruction from a single RGB image, we adopt LHM~\cite{qiu2025lhm}, which generalizes to in-the-wild inputs and enables real-time, pose-controllable avatars. Given an image (optionally with SMPL-X pose or a motion sequence), LHM extracts SMPL-X anchors and 2D visual tokens via a frozen vision encoder, fuses them in a multimodal transformer, and decodes per-Gaussian parameters (position, anisotropic scale, rotation, opacity, spherical-harmonic appearance). The resulting canonical 3D Gaussian representation preserves facial identity, clothing geometry, and texture, and can be skinned with linear blend skinning for photorealistic rendering and seamless animation.

\noindent \textbf{Single-View Scene Reconstruction.}
We utilize VistaDream ~\cite{wang2024vistadream} to reconstruct our 3D scene, which completes 3D scene from a single input image by progressively expanding and refining its content. It first expands the field of view through synthetic zoom-out and fills missing areas using diffusion-based inpainting guided by scene-specific text prompts, producing a complete RGB image and a monocular depth map to form a coarse 3D Gaussian scaffold. The method then warps this scaffold to novel viewpoints along the target camera trajectory, inpaints unseen regions, re-estimates depth, and incrementally augments the Gaussian field, improving coverage with each iteration. To achieve strict view-to-view consistency, VistaDream renders multiple viewpoints from the evolving Gaussian scene and applies view fusion inpainting, where diffusion-based denoising is blended with scaffold renderings to enforce cross-view coherence. The refined multi-view results are finally used to fine-tune the Gaussian field, yielding a high-fidelity, geometrically aligned 3D scene renderable from arbitrary novel viewpoints.

\subsection{Style Transfer Module}
We employ Stable Diffusion~\cite{rombach2022high} equipped with IP‑Adapter ~\cite{ye2023ip} to align the appearance of the human image $I_h$ with the style of the background scene image $I_s$. Specifically, the style reference image $I_s$ is fed into the IP-Adapter branch of Stable Diffusion, while the subject image $I_h$ is provided as the primary input for stylization. The IP‑Adapter modulates the diffusion process to inject the background’s style into the human rendering without altering its geometry or pose. This process produces a style-consistent human representation that can be seamlessly integrated into the reconstructed scene. Then we feed the style transfered  and the driven video $V_h$ into the the Large Animatable Human Model (LHM)~\cite{qiu2025lhm} to generate the 4D human sequence.

With style-consistent human and scene images, we reconstruct the 4D human sequence and the 3D scene using LHM~\cite{qiu2025lhm} and VistaDream~\cite{wang2024vistadream}, respectively. Specifically, we feed the human image $I_h$ and driving video $V_h$ into LHM to obtain a 4D human motion sequence that follows the motion in the driving video. Unlike VistaDream, which directly reconstructs the entire scene along a predefined camera trajectory, we only use the depth estimation part of VistaDream and obtain a sparse 3D representation by projecting scene pixels into 3D space. This enables a more accurate estimation of the 4D human sequence’s position within the scene. Once the human position is determined, performing inpainting and reconstruction along the camera trajectory prevents overlap between the 4D human Gaussians and the 3D scene Gaussians.

\subsection{Object Placement Module}
Existing object placement methods~\cite{gao2025object} operate exclusively in the two‐dimensional image domain, where a network predicts a bounding box $\mathbf{B}=(x,y,w,h)$, where $(x,y)$ denotes the top‐left corner and $(w,h)$ its width and height in the image $I_h$. Our task, however, requires estimating a plausible three‐dimensional location for an animatable 4D Gaussian human within a reconstructed 3D Gaussian scene, a problem that cannot be directly solved by these 2D‐focused models.

The set of $N$ uniformly spaced points along the bottom edge of this box is denoted by $\mathcal{E}=\{(u_n,v_n)\}_{n=1}^N$, and the depth map of the reconstructed Gaussian scene is given by $D(u,v)$. The camera intrinsic matrix is defined as
\begin{eqnarray}
K = \begin{bmatrix}
f_x & 0   & c_x\\
0   & f_y & c_y\\
0   & 0   & 1
\end{bmatrix},
\end{eqnarray}
and the resulting three‐dimensional insertion point is $\mathbf{p}^*=(X^*,Y^*,Z^*)$ in world coordinates.

First, an off‐the‐shelf 2D placement network is applied to predict the bounding box $\mathbf{B}=(x,y,w,h)$ for the human in the image $I$. Next, we sample $N$ points along the bottom edge of this box by setting 
\begin{eqnarray}
u_n = x + \frac{n}{N}\,w,\quad v_n = y + h,\quad n=1,\dots,N,
\end{eqnarray}
thereby defining $\mathcal{E}$. To estimate the depth coordinate $Z^*$, we compute the average of the depth map values at these sampled points,
\begin{eqnarray}
Z^* = \frac{1}{N}\sum_{n=1}^N D(u_n,v_n),
\end{eqnarray}
which mitigates noise from single‐pixel errors by aggregating information across the box baseline. We then select the center pixel of the bottom edge,
\begin{eqnarray}
(u_c,v_c) = \Bigl(x+\tfrac{w}{2},\,y+h\Bigr),
\end{eqnarray}
and back‐project it into 3D space using the inverse intrinsics:
\begin{eqnarray}
\mathbf{p}^* = Z^*\,K^{-1}
\begin{bmatrix}
u_c\\
v_c\\
1
\end{bmatrix}
= 
\begin{bmatrix}
X^*\\
Y^*\\
Z^*
\end{bmatrix}.
\end{eqnarray}
Finally, the 4D Gaussian human is positioned at $\mathbf{p}^*$ with a spatial covariance chosen so that its projection under the camera model matches the original 2D size $(w,h)$. To avoid penetration of the 4D Gaussian human into the reconstructed 3D Gaussian scene during motion, we enforce a collision-free constraint in world coordinates. Let $\mathcal{I}(\mathbf{q})$ be an occupancy indicator of the background Gaussian field, where $\mathcal{I}(\mathbf{q})=1$ denotes that $\mathbf{q}$ lies in a high-density background region, and $\mathcal{I}(\mathbf{q})=0$ otherwise. If $\mathcal{I}(\mathbf{p}^*)=1$, we project $\mathbf{p}^*$ to the nearest free-space point:
\begin{eqnarray}
\mathbf{p}^* \leftarrow \arg\min_{\mathbf{q} \in \mathbb{R}^3, \ \mathcal{I}(\mathbf{q}) = 0} \| \mathbf{q} - \mathbf{p}^* \|^2.
\end{eqnarray}
Even with collision-free projection, small frame-to-frame jitters or local overlaps may occur, so we apply temporal smoothing to the sequence of insertion points $\{\mathbf{p}^*_t\}_{t=1}^T$:
\begin{eqnarray}
\mathbf{p}^*_t \leftarrow (1-\lambda)\,\mathbf{p}^*_t + \lambda\,\frac{\mathbf{p}^*_{t-1} + \mathbf{p}^*_{t+1}}{2},
\end{eqnarray}
where $\lambda \in [0,1]$ controls the smoothing strength. This step guarantees a natural, continuous trajectory without abrupt spatial shifts while maintaining collision-free placement.

\subsection{Joint Post‑Reconstruction}
Once the 4D human motion sequence is properly inserted into the 3D scene, we follow the VistaDream\cite{wang2024vistadream} approach to reconstruct the fused scene along the predefined camera trajectory. To obtain higher-quality videos, after merging the reconstructed 3D scene with the 4D human and rendering along the target camera trajectory, we further performed post-processing on the video. Specifically, we selected the background color that contrasts significantly with the synthesized scene and we automatically detect all remaining unfilled cavities visible in the reconstructed color volume, resulting a binary mask that isolates these hole regions. To ensure seamless coverage of their boundaries, this mask is then expanded via a small radius morphological dilation. The dilated mask and the original reconstruction are passed together through our learned inpainting network, which hallucinates plausible content for each masked region by leveraging the surrounding geometry and appearance cues. Finally, the newly inpainted output is composited back into the 3D model, yielding a reconstruction with substantially reduced artifacts and holes while maintaining geometric fidelity and photorealistic detail.

\section{Experiments}
\subsection{Experimental setup}
\label{sec:experiment_setup}
We evaluate on 83 copyright-free single-view photos (indoor/outdoor, real and simulated) paired with 83 4D-human animation clips generated by LHM~\cite{qiu2025lhm}; each human instance is matched to a complementary scene, and six camera trajectories are designed to probe camera-controllable scene generation. Following prior work~\cite{wang2024vistadream}, we assess rendered frames under different camera paths with the VLM LLaVA~\cite{liu2023visual}, reporting LLaVA-IQA on noise suppression, edge sharpness, structural consistency, detail, and overall perceptual quality. As no existing method reconstructs camera-controllable 3D backgrounds for 4D humans, we compare against 3DitScene~\cite{zhang20243ditscene}—a language-guided disentangled Gaussian-splatting framework enabling single-image 3D scene generation and novel-view synthesis—and two generative view-synthesis methods, SEVA~\cite{zhou2025stable} (a generalist diffusion model for novel views given arbitrary inputs and target cameras) and DimensionX~\cite{sun2024dimensionx} (photorealistic 3D/4D scene generation from a single image via video diffusion).

\subsection{Evaluation results}
\begin{table}[t]
\begin{center}
\resizebox{\linewidth}{!}{
\begin{tabular}{lccccc} 
\toprule
Method & Noise-Free $\uparrow$ & Edge $\uparrow$ & Structure $\uparrow$ & Detail $\uparrow$ & Quality $\uparrow$ \\
\midrule
3DitScene~\cite{zhang20243ditscene}      & 0.696 & 0.262 & 0.533 & 0.645 & 0.538 \\
SEVA~\cite{zhou2025stable}              & 0.432 & 0.045 & 0.291 & 0.429 & 0.312 \\
DimensionX~\cite{sun2024dimensionx}     & 0.759 & 0.135 & 0.392 & 0.817 & 0.559 \\
Ours                                    & $\mathbf{0.980}$ & $\mathbf{0.344}$ & $\mathbf{0.636}$ & $\mathbf{0.979}$ & $\mathbf{0.759}$ \\
\bottomrule
\end{tabular}
}
\caption{LLaVA-IQA scores on synthetic video frames. AnimateScene ranks highest on five metrics.}
\label{tab:llava_score}
\end{center}
\end{table}
\begin{table}[t]
\begin{center}
\resizebox{\linewidth}{!}{
\begin{tabular}{cccccc}
\toprule & & 3DitScene~\cite{zhang20243ditscene} & SEVA~\cite{zhou2025stable}  & DimensionX~\cite{sun2024dimensionx}  & Ours \\
\midrule 
\multirow{2}{*}{Consistency} 
 & Human   & 17.26 & 7.84 & 18.91 & \textbf{55.99} \\
 & GPT4-v  & 11.37 & 4.68 & 22.05 & \textbf{61.90} \\
\midrule 
\multirow{2}{*}{Quality} 
 & Human   & 20.42 & 3.77 & 24.31 & \textbf{51.50} \\
 & GPT4-v  & 17.68 & 6.93 & 28.47 & \textbf{46.92} \\
\bottomrule
\end{tabular}
}
\caption{Human and GPT-4V preference shares on video quality and geometric consistency across methods, with AnimateScene achieving the highest votes.}
\label{tab:user_study}
\end{center}
\end{table}
 Since there is currently no method for reconstructing camera controllable 3D background scenes for 4D humans, we composite the baseline-generated controllable-view 3D scenes with the 4D human sequences frame by frame. Specifically, we render the human sequences according to the predefined camera trajectory, and then paste each rendered human frame onto the baseline-generated background according to positions provided by the object placement model, which ensures geometric consistency between the human and the scene in the final video. In Table~\ref{tab:llava_score}, we report the LLaVA scores across different dimensions, which showed that our method is capable of synthesizing high-quality scenes with controllable camera trajectories. 

To further evaluate the synthetic result, we conduct a user study to compare the rendered video by our method with the established baselines. We ask 20 users to vote for their preferred method based on video quality and geometric consistency across different viewpoints. Followed 3DitScene~\cite{zhang20243ditscene}, we also adopt GPT-4v as an additional criterion since it can evaluate 3D consistency and image quality well aligned with human preference. In Table~\ref{tab:user_study}, we report the user study result, which shows our method outperforms other baselines in terms of both video quality and geometric consistency.

\section{Conclusion} AnimateScene converts a single scene image, human image, motion clip, and camera path into a controllable 4D video. It first matches the actor’s appearance to the scene lighting, then embeds a depth-guided 3D avatar, and finally inpaints view-dependent gaps for seamless composition. Extensive quantitative tests and a user study show AnimateScene surpasses state-of-the-art baselines in noise suppression, edge sharpness, structural consistency, detail, perceptual quality, and geometric coherence.

\section{Acknowledgement}
The work was supported by the National Natural Science Foundation of China (Grant No. 62471287).

\bibliographystyle{IEEEbib}
\bibliography{refs}

\end{document}